\documentclass[sigconf, nonacm]{acmart}

\AtBeginDocument{%
  }

\usepackage{algorithm}
\usepackage{algorithmic}
\usepackage{booktabs}
\usepackage{balance}
\usepackage[htt]{hyphenat}
\usepackage{subcaption}
\usepackage{url}

\makeatletter
\@ifundefined{vldbdoi}{\newcommand{\vldbdoi}[1]{}}{}
\@ifundefined{vldbpages}{\newcommand{\vldbpages}[1]{}}{}
\@ifundefined{vldbvolume}{\newcommand{\vldbvolume}[1]{}}{}
\@ifundefined{vldbissue}{\newcommand{\vldbissue}[1]{}}{}
\@ifundefined{vldbyear}{\newcommand{\vldbyear}[1]{}}{}
\makeatother

\vldbdoi{XX.XX/XXX.XX}
\vldbpages{XXX-XXX}
\vldbvolume{19}
\vldbissue{X}
\vldbyear{2026}

\begin{document}

%\title{ADP-MA: Towards Self-Driving Data Pipelines\\through Hierarchical Agent Orchestration}
\title{Autonomous Data Processing using Meta-Agents}

\author{Udayan Khurana}
\email{ukhurana@us.ibm.com}
\affiliation{
  \institution{IBM T.J. Watson Research Center}
  \city{Yorktown Heights}
  \state{NY}
  \country{USA}
}

\renewcommand{\shortauthors}{Udayan Khurana}

\begin{abstract}
Building data processing pipelines for modern analytics is a labor-intensive and error-prone process, involving data profiling, schema alignment, multi-source integration, and domain-specific transformations. Large language models (LLMs) can generate individual code snippets, but autonomously building \emph{reliable} multi-stage pipelines raises additional challenges: managing cascading failures across stages, validating intermediate data products, and controlling cost under uncertain execution.
We present \textbf{ADP-MA}, a system for autonomous data pipeline construction based on \emph{hierarchical meta-agent orchestration}. Three persistent meta-agents handle planning and oversight while dynamically creating task-specific execution agents. The system provides several reliability mechanisms: progressive sampling for cost-aware validation, two-level backtracking for recovery from both implementation and planning errors, schema contracts for stage-level correctness, and rule-based monitoring that detects silent data corruption without additional model calls.
We evaluate ADP-MA on four benchmarks spanning 556 tasks in data science, scientific workflows, code generation, and database querying. ADP-MA outperforms all published single-agent baselines across multiple LLM backends. Ablation and scalability studies show that the architectural mechanisms contribute gains independent of which LLM is used.
\end{abstract}




\maketitle

%%% ================================================================
\section{Introduction}
\label{sec:intro}

Data processing pipelines are central to modern analytics, from business intelligence to scientific research and public-sector decision-making.
Yet building these pipelines remains a predominantly human-intensive process. A single pipeline may require data profiling, schema alignment, multi-source integration, domain-specific transformations, and quality validation---each demanding both programming skill and domain expertise.
The human attention involved in overseeing this process significantly influences the cost of analytics, and as organizations accumulate data faster than they can hire specialists, the gap between available data and actionable insight continues to widen.

Recent advances in large language models (LLMs) have produced coding assistants that generate data-processing snippets effectively. However, there is no well-defined basis for moving from snippet generation to \emph{autonomously orchestrating} a multi-stage pipeline. A working pipeline must handle schema mismatches across stages, recover from cascading failures, detect silent data corruption (e.g., a filter that drops 95\% of rows without raising an exception), and adapt to data characteristics discovered only at runtime.

This is fundamentally a \emph{data systems} problem. Traditional workflow engines (Airflow, Prefect) require users to manually specify DAGs. Query optimizers decompose declarative queries into physical plans but operate over fixed schemas. Self-driving databases tune knobs and indexes but do not construct pipelines from scratch. Autonomous pipeline construction requires a new abstraction that combines adaptive planning, runtime validation, and self-monitoring over underspecified natural-language intent rather than well-formed SQL.

Existing agentic approaches address parts of this challenge, but each has shortcomings that prevent it from solving the full problem. AutoKaggle~\cite{autokaggle} employs five agents but relies on predefined phase sequences. smolagents~\cite{smolagents} uses frontier LLMs in a flat iterative loop, achieving strong results through extended compute budgets but lacking plan-level recovery or continuous monitoring. DS-GURU~\cite{kramabench} and other single-agent baselines~\cite{dataagentsurvey} lack hierarchical orchestration entirely. AutoDCWorkflow~\cite{autodcworkflow} focuses on a single pipeline stage (data cleaning).

We argue that \emph{architecture matters as much as the choice of LLM} for autonomous data processing. Rather than relying solely on longer iterative loops, we adapt classical systems principles---validation, monitoring, contracts, and staged execution---to the setting where pipeline logic is synthesized dynamically. The central idea is to separate \emph{strategic orchestration} from \emph{task execution}, much as database systems separate query optimization from physical execution.

We present \textbf{ADP-MA} (Autonomous Data Processing using Meta-Agents), a system that constructs and executes data pipelines through \emph{hierarchical meta-agent orchestration}. A fixed set of meta-agents handles planning, monitoring, and recovery, while ephemeral ground agents generate and execute task-specific code in sandboxed environments. This separation enables reliability mechanisms that are hard to achieve in flat or single-agent architectures.

This paper makes the following contributions:

\begin{enumerate}
    \item \textbf{Problem formulation and system architecture for autonomous data pipelines.} We formalize autonomous pipeline construction as a systems problem requiring reliable orchestration of planning, code generation, execution, and validation under uncertainty. We introduce \emph{hierarchical meta-agent orchestration}, which separates strategic planning from task execution and enables reusable orchestration across heterogeneous data tasks (\S\ref{sec:architecture}).

    \item \textbf{Reliability mechanisms for LLM-driven data processing.} We design and implement mechanisms that provide robustness: \emph{progressive sampling} for cost-aware validation and early error detection, \emph{two-level backtracking} that recovers from both implementation failures and flawed pipeline decompositions, \emph{dynamically generated schema contracts} that enforce stage-level data invariants, and \emph{rule-based monitoring} that detects silent data corruption without additional LLM calls. Ablation studies quantify the individual and combined impact of these mechanisms (\S\ref{sec:implementation},~\S\ref{sec:evaluation}).

    \item \textbf{Comprehensive evaluation across diverse pipeline workloads.} We evaluate ADP-MA on four benchmarks comprising 556 tasks across data science, scientific workflows, code generation, and database querying, with up to five LLM families. Results show consistent improvements over single-agent baselines and competitive performance with multi-agent systems that use extended runtimes. A variance analysis quantifies run-to-run stability, finding 78\% of tasks deterministic with overall pass rates stable within $\pm$4pp (\S\ref{sec:evaluation}).

    \item \textbf{A deployable system and reproducible artifact.} We present a complete implementation with sandboxed execution, pluggable domain knowledge packs, and an interactive interface supporting real-time monitoring and replay. Source code, evaluation scripts, and all 12 domain knowledge packs will be publicly released.\footnote{Repository: \url{https://github.com/udayankhurana/adp-ma} (to be made public upon acceptance).}
\end{enumerate}

%%% ================================================================
\section{Related Work}
\label{sec:related}

\paragraph{Workflow engines and self-driving databases.}
Traditional workflow orchestrators (Airflow, Prefect) provide scheduling and monitoring but require users to manually specify DAGs. Query optimizers decompose SQL into physical plans but operate over fixed schemas. Self-driving databases automate tuning but do not construct pipelines. ADP-MA draws on all three: adaptive planning analogous to query optimization, integrity-constraint-like schema contracts, and self-driving monitoring---applied to constructing pipelines from underspecified natural-language intent.

\paragraph{Multi-agent data processing systems.}
AutoKaggle~\cite{autokaggle} employs five agents (Reader, Planner, Developer, Reviewer, Summarizer) across six fixed phases, achieving strong results on Kaggle competitions. However, its predefined phase sequences limit adaptability to novel task structures. smolagents~\cite{smolagents} adopts a Deep Research strategy where a frontier LLM (Claude-3.7 or GPT-o3) iteratively explores solution paths. While achieving 50\% on KramaBench~\cite{kramabench}, it relies on a flat agent loop without plan-level reasoning, cross-phase coordination, or monitoring. AutoPrep~\cite{autoprep} uses a multi-agent framework with planner, programmer, and executor agents for question-aware data preparation, demonstrating the value of decomposing data tasks across specialized agents. Prompt2DAG~\cite{prompt2dag} transforms natural-language specifications into executable Airflow DAGs, addressing pipeline generation but not runtime monitoring or adaptive recovery. Recent surveys~\cite{dataagentsurvey,dsagentsurvey} provide comprehensive taxonomies of LLM-based data science agents. ADP-MA extends this landscape with hierarchical meta-agent orchestration, two-level backtracking, and schema contracts.

\paragraph{Data cleaning and preparation.}
AutoDCWorkflow~\cite{autodcworkflow} generates data cleaning workflows via LLMs and OpenRefine operations. AutoClean~\cite{autoclean} uses LLM teams to generate cleaning rules from limited samples. CleanAgent~\cite{cleanagent} automates data standardization with specialized agents for column-type annotation, code generation, and execution. Research on cleaning--AutoML interaction~\cite{datacleaning_automl} shows that cleaning decisions should be optimized jointly with pipeline hyperparameters. Auto-Validate~\cite{autovalidate} programs data quality constraints from pipeline execution history, addressing recurring validation needs. ADP-MA treats data preparation as one adaptively configured phase within a broader pipeline, and its schema contracts provide column-level validation analogous to data contracts but generated dynamically per task.

\paragraph{LLMs for data management.}
The intersection of LLMs and data management has become a major research direction~\cite{data_llm}. Text-to-SQL systems~\cite{text2sql_benchmark,chess} demonstrate multi-agent approaches to database querying, with CHESS employing information retrieval, schema selection, and unit testing agents. Panda~\cite{panda} applies LLM agents to database performance debugging with grounding and verification mechanisms. ADP-MA complements these systems by targeting the broader pipeline construction problem beyond single-query generation.

\paragraph{Benchmarks.}
KramaBench~\cite{kramabench} provides 105 multi-step scientific pipelines requiring data discovery, integration, and statistical reasoning. DA-Code~\cite{dacode} benchmarks complex code generation across wrangling, ML, and EDA tasks. DSEval~\cite{dseval} evaluates data science agents across 825 problems from textbooks, StackOverflow, LeetCode, and Kaggle. AgentBench~\cite{agentbench} evaluates LLMs as interactive agents including database interaction. InsightBench~\cite{insightbench} evaluates end-to-end data analytics with multi-step insight generation. These benchmarks collectively stress the capabilities that ADP-MA targets: multi-step orchestration, domain-specific reasoning, code generation, and error recovery.

\paragraph{Concurrent LLM-based data systems.}
Several recent systems explore the intersection of LLMs and data management from complementary angles. SagaLLM~\cite{sagallm} integrates the Saga transactional pattern with persistent memory and automated compensation logic to ensure consistency across distributed agent workflows; ADP-MA's hierarchical backtracking is functionally analogous to SagaLLM's compensation but targets the data plane through schema contracts rather than transactional atomicity. DocETL~\cite{docetl} optimizes complex document processing pipelines through agent-guided logical rewrites; ADP-MA shares the goal of accuracy through decomposition but provides end-to-end pipeline \emph{construction} rather than rewriting user-defined operations, and targets tabular data rather than documents. LOTUS~\cite{lotus} introduces semantic operators (filter, join, ranking) with statistical accuracy guarantees over a declarative interface; ADP-MA operates at a layer above, orchestrating the pipeline that could theoretically use semantic operators as primitives.

\paragraph{Self-evolving agents and adaptive strategies.}
Recent work on self-evolving agents~\cite{selfevolvingsurvey} explores systems that improve through experience. Deep research agents~\cite{deepresearchagents} tackle complex multi-turn tasks through adaptive planning and retrieval. BacktrackAgent~\cite{backtrackagent} introduces verifier-judger-reflector components for error detection and recovery in GUI agents, validating the importance of structured backtracking that ADP-MA applies to data pipelines. SPIRAL~\cite{spiral} introduces a critic concept for dense intermediate rewards via reflection, which ADP-MA adapts for confidence-based early stopping. UCT-based approaches~\cite{fastmcts,abmcts} and SWE-Search~\cite{swesearch} apply Monte Carlo Tree Search to code generation and software engineering, inspiring ADP-MA's adaptive sampling strategy.

%%% ================================================================
\section{System Architecture}
\label{sec:architecture}

\begin{figure*}[t]
\centering
    \includegraphics[width=0.85\textwidth]{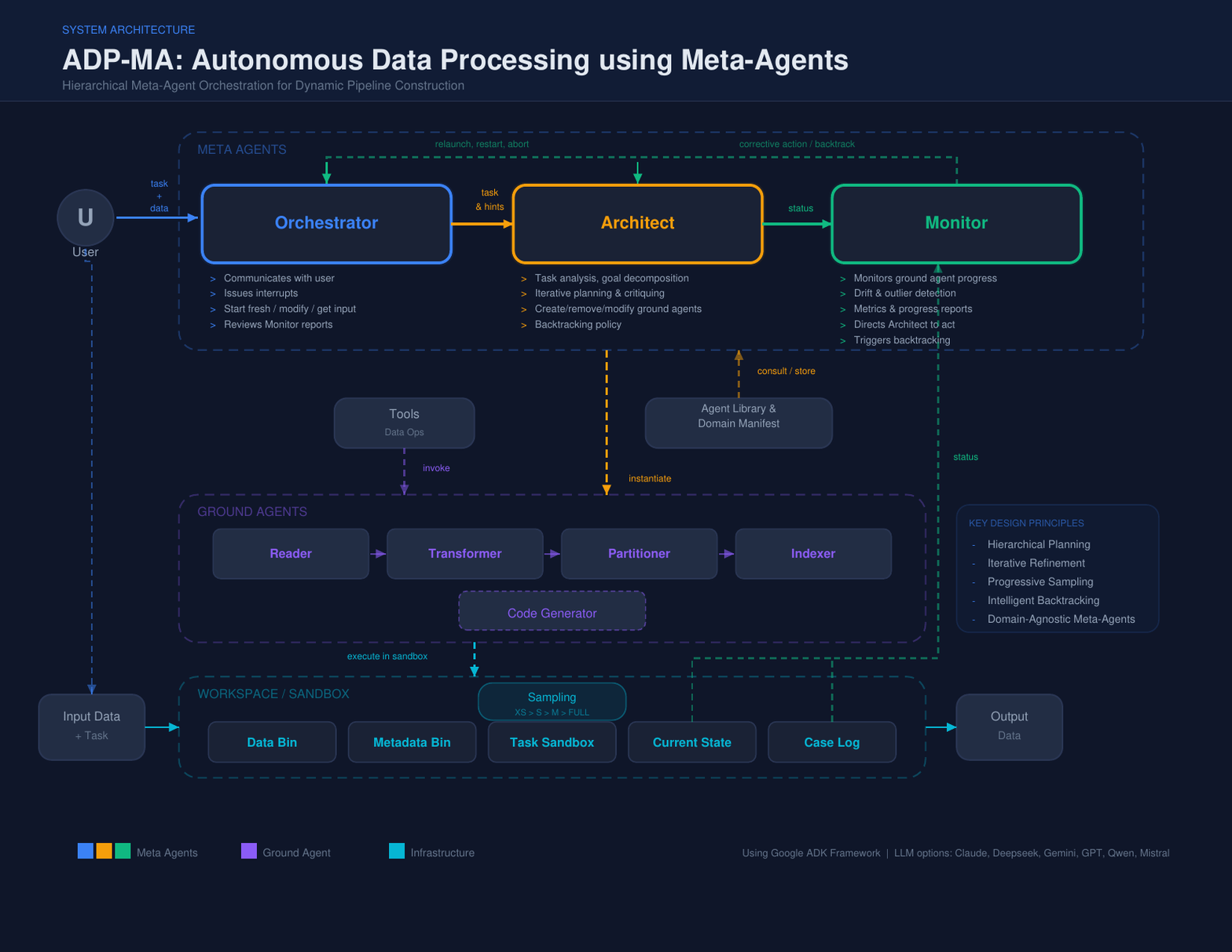}
    \caption{ADP-MA architecture. Three meta-agents (Orchestrator, Architect, Monitor) coordinate ground-level agents through a planning and execution workflow. Ground agents execute within process-level sandboxes with namespace isolation. Domain knowledge packs provide pluggable guidance via auto-detected keywords.}
    \label{fig:architecture}
\end{figure*}

ADP-MA has a hierarchical architecture with three layers: meta-agents for strategic orchestration, ground-level agents for task execution, and infrastructure for tool management, monitoring, and sandboxed execution. Figure~\ref{fig:architecture} shows the overall design.

\subsection{System Model and Assumptions}
\label{sec:system_model}

We model autonomous pipeline construction as follows. The system receives a task specification $T$ (natural-language goal) and input data $D = \{d_1, \ldots, d_k\}$ (one or more tabular files). The objective is to produce a correct output $O$ by constructing and executing a pipeline $\Pi = (s_1, \ldots, s_n)$ of data transformation stages, where each stage $s_i$ is a Python function operating on DataFrames.

The key challenges are: (1)~\emph{underspecification}---$T$ is natural language, not a formal query, so the system must infer missing semantics from data profiling; (2)~\emph{cascading failures}---an error in $s_i$ propagates silently to all downstream stages $s_{i+1}, \ldots, s_n$; (3)~\emph{non-determinism}---LLM-generated code is stochastic, so the same prompt may produce correct or incorrect implementations across runs; (4)~\emph{cost control}---each LLM call has monetary and latency cost, requiring the system to balance exploration against budget.

\paragraph{Failure model.}
Failures may occur at four levels: \emph{code failures} (syntax, runtime, type errors), \emph{semantic failures} (incorrect transformations), \emph{data-quality failures} (schema mismatch, silent row drops), and \emph{planning failures} (incorrect task decomposition). Table~\ref{tab:failure_coverage} maps each failure type to the mechanism that addresses it. The mechanisms are complementary: each covers a distinct failure class, and no single mechanism subsumes the others. We do not guarantee semantic correctness without ground truth; the system maximizes reliability using structural signals and iterative refinement.

\begin{table}[t]
\caption{Failure types and the ADP-MA mechanism that addresses each. Each mechanism covers a distinct class; their composition provides layered defense.}
\label{tab:failure_coverage}
\centering
\small
\begin{tabular}{p{0.17\columnwidth} p{0.27\columnwidth} p{0.42\columnwidth}}
\toprule
\textbf{Failure type} & \textbf{Caught by} & \textbf{Example} \\
\midrule
Code & Sandbox + local retry & ImportError, NameError, type mismatch \\
\addlinespace
Semantic & Progressive sampling + schema contracts & Wrong aggregation, missing filter, column mismatch \\
\addlinespace
Data-quality & Rule-based Monitor & Row explosion, null-rate spike, silent row drop \\
\addlinespace
Planning & Two-level backtracking & Single-phase plan for a multi-source join \\
\bottomrule
\end{tabular}
\end{table}

\paragraph{Assumptions and scope.}
We assume: (a)~input data fits in memory (sampling handles larger inputs); (b)~an LLM API is available for Python code generation; (c)~tasks decompose into DataFrame transformation pipelines; (d)~no oracle feedback at execution time. We target tabular data processing and do not address real-time streaming, distributed execution, or formal verification of semantic correctness.

\subsection{Design Principles}

The architecture is built around five principles:
(1)~\textbf{Hierarchical planning}---decompose tasks into phases before generating code, avoiding context degradation~\cite{anthropiccontext};
(2)~\textbf{Iterative refinement}---severity-gated critique cycles at plan-level and phase-level;
(3)~\textbf{Progressive sampling}---validate code on 1\%, 5\%, 25\% data subsets before full execution;
(4)~\textbf{Intelligent backtracking}---local code refinement and global plan revision;
(5)~\textbf{Domain-agnostic meta-agents}---meta-agents reason about structure, not domain expertise, which is injected via pluggable packs.

\paragraph{Analogy to query optimization.}
These principles adapt classical database systems ideas to a setting where the pipeline logic itself is synthesized from natural language. Table~\ref{tab:db_analogy} makes the mapping explicit. The Orchestrator produces a \emph{logical plan} (what phases to execute); the Architect selects a \emph{physical plan} (which agent types to instantiate, what code strategy to use). Progressive sampling plays the role of cost-based optimization: the system tests cheap before committing. Two-level backtracking corresponds to re-optimization when runtime statistics (e.g., cardinality estimates) invalidate the chosen plan. Schema contracts enforce inter-operator type invariants, and the Monitor serves as a runtime watchdog. The key difference from traditional query processing is that the ``query'' is underspecified---natural language rather than SQL---so the planner must infer missing semantics from data profiling. This analogy is not cosmetic: it grounds each mechanism in a well-understood systems principle and clarifies why the specific combination of mechanisms produces gains that no single mechanism achieves alone.

\begin{table}[t]
\caption{Mapping from classical database systems concepts to ADP-MA mechanisms.}
\label{tab:db_analogy}
\centering
\small
\begin{tabular}{p{0.38\columnwidth} p{0.52\columnwidth}}
\toprule
\textbf{Database concept} & \textbf{ADP-MA mechanism} \\
\midrule
Logical plan & Orchestrator's phase decomposition \\
\addlinespace
Physical plan & Architect's agent assignments + code strategy \\
\addlinespace
Cost-based optimization & Progressive sampling (test on 1\% before committing) \\
\addlinespace
Re-optimization on cardinality misestimate & Two-level backtracking (plan was wrong, not just code) \\
\addlinespace
Inter-operator type checking & Schema contracts (column-level invariants) \\
\addlinespace
Integrity constraints & Dynamically generated per-task contracts \\
\addlinespace
Runtime query watchdog & Rule-based Monitor (0 LLM calls) \\
\bottomrule
\end{tabular}
\end{table}

\subsection{Meta-Agent Layer}

Three meta-agents persist unchanged across tasks and communicate through shared pipeline state. Following the query optimization analogy (Table~\ref{tab:db_analogy}), the Orchestrator produces a logical plan while the Architect selects a physical plan~\cite{selinger1979,graefe1995cascades}.

\noindent\textbf{Orchestrator.} Serves as the primary interface between users and the system. It analyzes task specifications, profiles input data (schema inference, statistics, natural-language summary), and decomposes goals into structured phases. The Orchestrator operates at a strategic level: it does not generate code or manipulate data directly, but produces the data understanding context that informs all downstream decisions.

\noindent\textbf{Architect.} Translates high-level plans into concrete implementations. It creates ground-level agents with appropriate types, tools, and schema contracts; implements a two-level critique mechanism (plan-level and expansion-level); and makes backtracking decisions when execution fails. The Architect maintains a library of reusable agent definitions for rapid pipeline composition.

\noindent\textbf{Monitor.} Provides continuous, rule-based oversight of pipeline execution. It tracks progress across all active ground agents, detects anomalies (row-count explosions, null-rate spikes, cost overruns) via configurable thresholds, and issues structured verdicts (\textsc{continue}, \textsc{warn}, \textsc{pause}, \textsc{abort}, \textsc{retry}) without incurring any LLM calls. The Monitor also supports cross-run learning by loading failure patterns from prior cases.

None of the three meta-agents encode domain-specific logic. Rather than hard-coding expertise, meta-agents assess the nature of the problem and instantiate appropriately specialized ground agents. We chose three meta-agents after evaluating alternatives. A two-agent design (planner + executor) conflates monitoring with execution, losing independent oversight. A four-agent design (splitting Architect into Planner and Code Architect) adds coordination overhead without measurable quality gains in our experiments. Three agents provide the minimum separation of concerns---planning, implementation, and oversight---needed for two-level backtracking.

\subsection{Ground-Level Agent Ecosystem}

Ground-level agents are ephemeral and task-specific. Each agent is instantiated by the Architect with a specified type (DataProcessor, Aggregator, AnswerFormatter, Joiner, etc.), an objective, implementation hints, and a schema contract defining expected inputs and outputs. Agents generate Python code, execute it within a process-level sandbox (Section~\ref{sec:sandbox}), and report structured results.

Table~\ref{tab:meta_vs_ground} contrasts the key characteristics of meta-agents and ground-level agents.

\begin{table}[t]
\caption{Contrasting characteristics of meta-agents and ground-level agents.}
\label{tab:meta_vs_ground}
\centering
\small
\begin{tabular}{p{0.16\columnwidth} p{0.36\columnwidth} p{0.36\columnwidth}}
\toprule
\textbf{Aspect} & \textbf{Meta-Agents} & \textbf{Ground Agents} \\
\midrule
Role & Planning, orchestration, oversight & Task-specific code execution \\
\addlinespace
Lifespan & Persistent across pipeline & Ephemeral; created per task \\
\addlinespace
Data access & Indirect (summaries, metadata) & Direct (read, transform, write) \\
\addlinespace
Code gen. & None & Generate \& execute in sandbox \\
\addlinespace
Cardinality & Fixed (3) & Variable; on demand \\
\addlinespace
Failure & Decide backtracking scope & Local code refinement \\
\bottomrule
\end{tabular}
\end{table}

\subsection{Operational Workflow}

Pipeline construction follows six stages, summarized in Algorithm~\ref{alg:workflow}:

\begin{algorithm}[t]
\caption{ADP-MA Pipeline Construction}
\label{alg:workflow}
\begin{algorithmic}[1]
\REQUIRE Task specification $T$, Input data $D$
\ENSURE Processed output $O$
\STATE \textbf{// Stage 1: Data Understanding}
\STATE $S \leftarrow$ \textbf{Orchestrator} profiles $D$ (schema, stats, summary)
\STATE \textbf{// Stage 2: High-Level Planning}
\STATE $P = \{p_1, \ldots, p_n\} \leftarrow$ \textbf{Architect} plans phases from $T, S$
\STATE \textbf{// Stage 3: Critique and Refinement}
\REPEAT
    \STATE $C \leftarrow$ \textbf{Architect} critiques $P$
    \IF{$C.\mathrm{severity} \geq$ \textsc{major}}
        \STATE $P \leftarrow$ \textbf{Architect} revises plan using $C$
    \ENDIF
\UNTIL{$C.\mathrm{severity} \leq$ \textsc{minor} \OR max iterations}
\STATE \textbf{// Stage 4: Phase Expansion}
\FOR{each phase $p_i \in P$}
    \STATE $E_i \leftarrow$ \textbf{Architect} expands $p_i$ into substeps
\ENDFOR
\STATE \textbf{// Stage 5: Ground-Agent Execution}
\FOR{each substep $e \in E_1 \cup \cdots \cup E_n$ \textbf{(per strategy)}}
    \FOR{$\ell \in$ \{XS, S, M, FULL\}}
        \STATE Generate code via LLM; execute on $\ell$-sample
        \IF{execution fails}
            \STATE Refine with error context; retry (up to budget)
        \ENDIF
    \ENDFOR
    \STATE \textbf{Monitor} evaluates output; issue verdict
\ENDFOR
\STATE \textbf{// Stage 6: Finalization}
\STATE Assemble pipeline; write case documentation
\RETURN Final processed output $O$
\end{algorithmic}
\end{algorithm}

The framework supports separate LLM configurations for planning (meta-agents) and coding (ground agents), allowing users to assign different models to different roles. Supported providers include Anthropic, Google, OpenAI, DeepSeek, Ollama and any other providers that use these style of APIs.

%%% ================================================================
\section{System Implementation}
\label{sec:implementation}

This section describes the mechanisms that provide correctness guarantees independent of the underlying LLM.

\subsection{Progressive Sampling}
\label{sec:progressive}

Each ground agent is validated through an escalating sequence of sample sizes (Table~\ref{tab:sampling}), inspired by adaptive query processing~\cite{avnur2000eddies}. Rather than committing to a fixed execution plan, the system evaluates intermediate results at each level and adapts its strategy (refine, scale up, or restart). Execution begins at XS (10 rows). On success, the agent promotes to the next level; on failure, the coding LLM receives the error traceback and generates a revised implementation at the same level. A maximum refinement budget (default 3) caps retry cost.

\begin{table}[t]
\caption{Progressive sampling levels.}
\label{tab:sampling}
\centering
\small
\begin{tabular}{lrl}
\toprule
\textbf{Level} & \textbf{Rows} & \textbf{Purpose} \\
\midrule
XS & 10 & Syntax and basic logic validation \\
S  & 100 & Functional correctness \\
M  & 1000 & Performance and edge-case coverage \\
FULL & All & Production execution \\
\bottomrule
\end{tabular}
\end{table}

\paragraph{UCT-based adaptive sampling.}
As an alternative to linear progression, ADP-MA supports UCT-based adaptive sampling inspired by MCTS for code generation~\cite{fastmcts,abmcts,spiral}. At each decision point, the strategy selects among four actions using the UCT formula:
\begin{equation}
\text{UCT}(i) = \frac{w_i}{n_i} + c \cdot \sqrt{\frac{\ln N_i}{n_i}}
\end{equation}
where $w_i$ is cumulative reward, $n_i$ is visit count, $N_i$ is parent visits, and $c$ is the exploration constant (default $\sqrt{2}$). The four actions are: \textbf{SCALE\_UP} (promote to next level), \textbf{REFINE} (new code variant at current level), \textbf{RESTART} (abandon approach, regenerate from XS), and \textbf{COMMIT} (accept result, skip remaining levels). The exploration constant adapts by depth: $1.5c$ at XS, $1.2c$ at S, $c$ at M, $0.8c$ at FULL---encouraging diverse code variants when failure cost is low.

\paragraph{Confidence-based early stopping.}
Inspired by SPIRAL's critic~\cite{spiral}, a \emph{ValidationCritic} evaluates cross-level output stability to determine whether full-scale execution can be skipped. The critic computes a composite confidence score as the product of four sub-scores: (1)~\emph{schema stability}---identical column names and dtypes across levels; (2)~\emph{row-count stability}---consistent proportional or fixed output pattern; (3)~\emph{time scaling}---at most linear growth with input size; (4)~\emph{level depth}---three validated levels yield 0.95, two yield only 0.60. When the composite score exceeds threshold $\tau = 0.92$, the strategy returns \textsc{commit}. The critic operates without LLM calls, inspecting only execution metadata.

\subsection{Two-Level Critique and Backtracking}
\label{sec:critique}

The critique mechanism serves as the quality gate before code generation:

\noindent\textbf{Level~1 (Plan Critique).} Evaluates the sequence of phases against the task context. Key checks: phase ordering respects data dependencies, every goal aspect is addressed, no phase introduces out-of-scope work, and phase count is minimal. Severity levels: \textsc{none}, \textsc{minor}, \textsc{major}, \textsc{critical}.

\noindent\textbf{Level~2 (Expansion Critique).} Evaluates each phase's substep decomposition. Checks that agent types are appropriate, input columns required by each substep are available from predecessors, and substep count stays within budget (1--3 per phase).

Both levels share exit logic parameterized by \texttt{CritiqueLoopConfig}: maximum iterations (default 10), exit severity threshold, optional convergence detection, and optional dual-judge mode.

When ground-agent execution fails, two-level backtracking provides recovery:
\begin{itemize}
    \item \textbf{Phase-level:} Discard failing substep outputs, revert to the prior successful state, and re-expand the current phase with different agent types or hints.
    \item \textbf{Plan-level:} When phase-level backtracking fails or the Monitor determines a fundamental plan flaw, backtrack to the planning stage with accumulated error evidence for a revised high-level plan.
\end{itemize}
Heuristic triggers route between levels: single substep failures initiate phase-level backtracking; repeated phase-level failures (threshold: 2) escalate to plan-level. Per-phase and global retry caps prevent unbounded loops.

\noindent\textbf{State management.}
Each pipeline maintains a case directory containing checkpointed outputs: the data-understanding profile, the current plan, and each phase's output DataFrame. Phase-level backtracking discards the failing phase's outputs and restores the predecessor phase's DataFrame as the active input. Plan-level backtracking resets the pipeline state to the data-understanding checkpoint---preserving schema profiles and statistics---while retaining the accumulated failure log so the Architect can avoid repeating the same strategy. Because ground agents run in isolated sandboxes and write only to their assigned output paths, partial results from discarded phases never contaminate the shared state.

\subsection{Schema Contracts}
\label{sec:contracts}

Each ground agent is governed by an \texttt{EnhancedSchemaContract} specifying: required input columns with type constraints, columns to add/preserve/remove, value constraints (range, enum, regex, uniqueness), row-count constraint relative to input (\textsc{same}, \textsc{less}, \textsc{greater}, \textsc{any}), and free-text postconditions. These contracts are the autonomous-pipeline analogue of integrity constraints in traditional DBMS~\cite{fangeerts2012}. Just as CHECK constraints, foreign keys, and NOT NULL declarations enforce data quality at the storage layer, schema contracts enforce structural invariants at the pipeline-stage layer. The difference is that contracts are \emph{dynamically generated per task} by the Architect, adapting to each pipeline's data-flow requirements rather than being statically defined at schema creation time. Contracts flow through the Architect's expansion prompts into ground-agent code-generation prompts. At execution time, violations are recorded with severity and affected column, enabling targeted refinement.

\subsection{Pipeline Monitoring}
\label{sec:monitor}

Monitoring is essential when LLM-generated code has unsupervised data access. In a flat agent loop, the LLM is both executor and judge---it cannot detect failures it does not recognize as failures. ADP-MA's separation of concerns enables an independent Monitor that is \emph{deterministic} (no stochasticity), \emph{zero-cost} (no LLM calls), and capable of catching failures that LLMs \emph{cannot self-detect}. A silent Cartesian-product explosion, for example, produces a valid DataFrame with plausible column names; no LLM examining its own output would flag it, but a row-count rule catches it immediately.

Four properties make monitoring a structural necessity: (1)~\emph{cascading failure prevention}---a silent row drop produces a valid DataFrame but corrupts all downstream agents; (2)~\emph{resource protection}---cost tracking prevents runaway LLM spending; (3)~\emph{cross-run learning}---failure patterns from prior runs become preemptive warnings; (4)~\emph{separation of judge from executor}---the Monitor's independence from the code-generating LLM ensures that oversight is not subject to the same failure modes as generation.

The Monitor evaluates system state after every ground-agent execution using rule-based heuristics (Table~\ref{tab:monitor}) without LLM calls.

\begin{table}[t]
\caption{Monitoring thresholds.}
\label{tab:monitor}
\centering
\small
\begin{tabular}{lll}
\toprule
\textbf{Check} & \textbf{Warning} & \textbf{Critical} \\
\midrule
Revision count & $\geq 2$ & $\geq 4$ \\
Row drop (\%) & $\geq 30\%$ & $\geq 90\%$ \\
Row growth (\%) & $\geq 500\%$ & --- \\
Null-rate increase & $\geq 20$ pp & --- \\
Agent wall-clock & $\geq 60$\,s & $\geq 300$\,s \\
Cost vs.\ budget & $\geq 80\%$ & $\geq 100\%$ \\
\bottomrule
\end{tabular}
\end{table}

Thresholds (Table~\ref{tab:monitor}) were tuned on a development set of 30 tasks across 6 domains: the row-drop warning at 30\% balances sensitivity (catching silent filters) against false alarms (legitimate aggregations); the critical threshold at 90\% was determined empirically as the point where downstream agents invariably produce wrong answers. A Failure Analysis Module extends the monitor with 14 regex-based error categories and domain-specific recovery hints, supporting both synchronous (per-execution) and asynchronous (cross-case) analysis.

\subsection{Code Execution Sandbox}
\label{sec:sandbox}

Each ground agent's code executes within a process-level sandbox providing namespace isolation and resource tracking:
(1)~A restricted namespace is constructed with pre-imported libraries (pandas, datetime) and input DataFrames (copied to prevent mutation).
(2)~Code is executed via Python's \texttt{exec}; stdout is captured; peak memory is tracked via \texttt{tracemalloc}.
(3)~The sandbox auto-detects functions matching the \texttt{stage\_*} naming convention and invokes them.
(4)~Structured results include success status, output DataFrame, execution time, memory usage, and error tracebacks.

\subsection{Domain Knowledge Packs}
\label{sec:domain_packs}

Twelve domain-specific knowledge packs span scientific, legal, financial, and analytical domains (astronomy, biomedical, environment, legal, archaeology, wildfire, ML, visualization, finance, sports, ecommerce, statistics). Each pack contains: (i)~keywords for automatic domain detection from task objectives and column names; (ii)~a domain-expert system prompt injected into ground-agent prompts; (iii)~planning guidance; (iv)~data format recipes; (v)~gotchas (e.g., ``BP dates use 1950 CE as Present''); (vi)~recommended tools. Multi-domain detection merges guidance when tasks span domains. Extending to a new domain requires writing a single knowledge pack file---no meta-agent changes. All 12 packs are included in the released repository as self-contained Python modules.

\subsection{Prompt Budget Management}
\label{sec:prompt_budget}

Hierarchical planning (\S\ref{sec:architecture}) prevents \emph{inter-call} context rot by separating strategic reasoning from code generation. However, \emph{intra-call} degradation remains: each prompt must convey schema metadata, domain guidance, pipeline context, and output contracts within a single context window. When datasets are wide---a 200-column DataFrame serializes to ${\sim}$20K tokens of column names and types alone---code quality suffers, particularly for models with smaller context budgets. We treat prompt construction as a resource allocation problem under a per-model token budget.

\noindent\textbf{Task-relevant projection.}
Similar to column projection in query optimization, schema metadata is projected to only the columns likely to be referenced in the generated code. The projection is \emph{goal-directed}: columns whose names share keywords with the task objective are retained preferentially, since LLMs attend more reliably to schema elements that echo the natural-language goal. The projection is a no-op for narrow datasets (${\leq}$30 columns), ensuring zero overhead on the common case.

\noindent\textbf{Priority-ordered prompt composition.}
Rather than treating the prompt as a monolithic string, we structure it as a composition of typed blocks with different marginal value to the LLM. Critical blocks (function contract, output schema, pipeline context) carry highest priority; auxiliary blocks (tool hints, merge-strategy templates) have diminishing returns. Under a fixed token budget---set per model based on empirical context-tolerance thresholds---blocks are assembled greedily by priority. Dropping a low-priority block (e.g., a tool-availability listing) has negligible impact on code correctness, whereas truncating a high-priority block (e.g., the output schema) almost always causes a downstream failure.

\noindent\textbf{Demand-driven guidance materialization.}
Domain knowledge packs (\S\ref{sec:domain_packs}) provide supplementary context---gotchas, data-format recipes, value-range checks---but injecting all guidance into every prompt is wasteful. Guidance about headerless-file parsing is irrelevant to an aggregation substep. We materialize guidance \emph{on demand}: each item is included only when its keywords overlap with the current substep objective, reducing prompt size while preserving coverage where it matters.

\noindent\textbf{Admission control and graceful degradation.}
Before each LLM call, the system estimates prompt size and applies admission control: a warning at 80\% of the model's budget and truncation at 100\%. Truncation preserves the prompt head (where the task specification lives) and the tail (where the return-format instruction appears), sacrificing the interior, which is empirically the region most tolerant to omission. Error tracebacks in refinement calls are similarly capped to the tail, where the actual exception appears, discarding redundant stack frames.

These four mechanisms complement progressive sampling (\S\ref{sec:progressive}) along an orthogonal axis: sampling controls \emph{data volume} seen during execution, while prompt budgeting controls \emph{metadata volume} seen during code generation.

\subsection{Execution Strategies}
\label{sec:strategies}

Three strategies govern how ground agents are dispatched:

\begin{itemize}
    \item \textbf{Centralized (default):} Serial execution. Each agent receives full accumulated context. Maximizes inter-agent information flow; best for tight data dependencies. This is the default because our evaluation shows autonomous execution fails on multi-agent plans (0/3 vs.\ 2/3 for centralized) due to contract-only context sharing (\S\ref{sec:scalability}).
    \item \textbf{Autonomous:} Parallel batches grouped by dependency depth via \texttt{asyncio.gather}. Contract-only context (no accumulated state). Best for independent stages.
    \item \textbf{Hybrid:} Phases execute sequentially (preserving inter-phase flow); agents within a phase execute in parallel. Balances reliability and throughput.
\end{itemize}

\subsection{Case Documentation}
\label{sec:casedocs}

Every run produces a self-contained case folder with all meta-agent decisions, monitor alerts, a time-stamped event stream for replay, generated code with revision history, and an assembled standalone pipeline. This ensures full reproducibility and enables the interactive replay interface.

%%% ================================================================
\section{Evaluation}
\label{sec:evaluation}

We evaluate ADP-MA on four benchmarks: DSEval~\cite{dseval} (data science tasks), KramaBench~\cite{kramabench} (multi-step scientific pipelines), DA-Code~\cite{dacode} (hard code generation), and AgentBench DBBench~\cite{agentbench} (database queries).

\paragraph{Methodology note.}
ADP-MA delegates all computationally intensive work---LLM inference, code generation, and iterative refinement---to remote model-hosting infrastructure, so differences in client-side hardware have negligible impact on task outcomes. Published results from competing systems, which similarly rely on hosted LLM endpoints, are therefore directly comparable regardless of the original authors' local setup. Given the substantial cost of full benchmark reproduction, we compare against results reported in the respective benchmark papers rather than re-running all baselines. The landscape of available models is expanding rapidly, making exhaustive evaluation across all configurations cost-prohibitive. We present results for three representative LLMs and leave the interpretation of cross-model comparisons partly to the reader.

\subsection{Experimental Setup}
\label{sec:setup}

\paragraph{Hardware and software environment.}
All experiments run on an Apple M1 Pro (10-core CPU, 16\,GB unified memory) under macOS~13.0 with Python~3.11. No GPU is used; all LLM inference is performed via remote API calls. The ADP-MA backend runs as a single FastAPI process with \texttt{asyncio} concurrency for parallel agent dispatch.

\paragraph{LLM configurations.}
Primary experiments use DeepSeek-3.2 (deepseek-chat) via the DeepSeek API with centralized execution strategy and a maximum of 3 code refinements per agent. To validate model-agnosticity, we additionally evaluate with Claude Sonnet~4.5 (Anthropic API) and GPT-4o (OpenAI API) under identical settings. All models use temperature~0 and seed-based determinism where supported. Planning and coding use the same model per run; no model mixing within a single pipeline.

\paragraph{Latency decomposition.}
Table~\ref{tab:latency} decomposes wall-clock time by pipeline stage across benchmarks. LLM API latency dominates (70--85\% of total time), with local computation (data profiling, code execution, monitoring) consuming the remainder. Planning is a fixed cost independent of data volume; coding scales with refinement count.

\begin{table}[t]
\caption{Latency decomposition by pipeline stage (DeepSeek-3.2, averaged across benchmark tasks).}
\label{tab:latency}
\centering
\small
\begin{tabular}{l@{\hspace{5pt}}r@{\hspace{5pt}}r@{\hspace{5pt}}r@{\hspace{5pt}}r}
\toprule
\textbf{Stage} & \textbf{DSEval} & \textbf{Krama} & \textbf{DA-Code} & \textbf{AgentB.} \\
\midrule
Data profiling & 1.2s & 2.8s & 1.5s & 1.0s \\
Plan + critique & 8.4s & 12.1s & 9.6s & 7.2s \\
Code generation & 5.1s & 18.3s & 14.2s & 8.5s \\
Execution & 0.8s & 3.4s & 2.1s & 1.8s \\
Monitoring & 0.1s & 0.2s & 0.1s & 0.1s \\
\midrule
\textbf{Total} & \textbf{15.6s} & \textbf{36.8s} & \textbf{27.5s} & \textbf{18.6s} \\
LLM \% of total & 87\% & 84\% & 87\% & 84\% \\
\bottomrule
\end{tabular}
\end{table}

\subsection{Cross-Benchmark Results}

Table~\ref{tab:cross_benchmark} summarizes ADP-MA's performance against published baselines.

\paragraph{Baseline methodology.} All ADP-MA results are from our own runs under the controlled environment described in \S\ref{sec:setup}. Baseline numbers for DSEval (CoML), KramaBench (DS-GURU, smolagents, OpenAI DR), DA-Code (DA-Agent), and AgentBench (GPT-4) are taken from the respective benchmark papers and leaderboards, which use different hardware and LLM budgets. We note this limitation: a fully controlled comparison would require re-implementing each baseline under identical conditions, which we leave to future work. Where possible, we compare against baselines using the \emph{same} underlying LLM (e.g., GPT-4o vs.\ GPT-4) to reduce confounds.

\begin{table*}[t]
\caption{Cross-benchmark comparison. ADP-MA outperforms all single-agent baselines on every benchmark. Best result per benchmark in \textbf{bold}. Avg LLM calls and duration shown for best ADP-MA configuration.}
\label{tab:cross_benchmark}
\centering
\small
\resizebox{\textwidth}{!}{%
\begin{tabular}{l@{\hspace{5pt}}r@{\hspace{5pt}}r@{\hspace{5pt}}r@{\hspace{5pt}}r@{\hspace{5pt}}r@{\hspace{5pt}}r@{\hspace{5pt}}r@{\hspace{5pt}}l@{\hspace{5pt}}l}
\toprule
& \multicolumn{5}{c}{\textbf{ADP-MA +}} & & & & \\
\cmidrule(lr){2-6}
\textbf{Benchmark} & \textbf{GPT-4o} & \textbf{GPT-5} & \textbf{DeepSeek-3.2} & \textbf{Sonnet 4.5} & \textbf{Gemini 2.5 Pro} & \textbf{Calls} & \textbf{Dur (s)} & \textbf{Best Single} & \textbf{Best Multi} \\
\midrule
DSEval (299) & 81.3\% & 83.6\% & 89.6\% & \textbf{90.6\%} & 86.6\% & 4.8 & 99 & 76.2\% (CoML) & --- \\
KramaBench (105) & 26.7\% & --- & 41.0\% & \textbf{44.8\%} & 38.5\% & 5.9 & 139 & 22.1\% (DS-GURU) & 50.0\% \\
DA-Code (52) & 32.7\% & 42.3\% & 48.1\% & 44.2\% & \textbf{50.0\%} & 3.7 & 553 & 23.4\% (GPT-4) & --- \\
AgentBench (100) & 56.0\% & 70.0\% & 61.0\% & \textbf{70.0\%} & 65.0\% & 4.4 & 276 & 32.0\% (GPT-4) & $\sim$52\% \\
\bottomrule
\end{tabular}}
\end{table*}

\subsection{DSEval Results}

DSEval comprises 299 tasks across five categories (Examples, Exercise, Kaggle, LeetCode, StackOverflow). Table~\ref{tab:dseval} presents results with five LLMs.

\begin{table*}[t]
\caption{DSEval results by category (five LLMs).}
\label{tab:dseval}
\centering
\small
\begin{tabular}{lrccccc}
\toprule
& & \multicolumn{5}{c}{\textbf{ADP-MA +}} \\
\cmidrule(l){3-7}
\textbf{Category} & \textbf{N} & \textbf{DeepSeek-3.2} & \textbf{Sonnet 4.5} & \textbf{GPT-4o} & \textbf{GPT-5} & \textbf{Gemini 2.5} \\
\midrule
Examples & 5 & \textbf{100\%} & \textbf{100\%} & \textbf{100\%} & \textbf{100\%} & 60.0\% \\
Exercise & 21 & \textbf{100\%} & \textbf{100\%} & 95.2\% & \textbf{100\%} & 95.2\% \\
Kaggle & 31 & \textbf{100\%} & \textbf{100\%} & 90.3\% & \textbf{100\%} & \textbf{100\%} \\
LeetCode & 40 & 95.0\% & 92.5\% & 87.5\% & 95.0\% & \textbf{97.5\%} \\
StackOverflow & 202 & 85.6\% & \textbf{87.6\%} & 76.7\% & 76.7\% & 82.2\% \\
\midrule
\textbf{Total} & \textbf{299} & \textbf{89.6\%} & \textbf{90.6\%} & \textbf{81.3\%} & \textbf{83.6\%} & \textbf{86.6\%} \\
\midrule
Avg LLM calls & & 10.9 & 4.8 & 4.3 & 5.1 & 4.2 \\
Avg duration (s) & & 94 & 99 & 114 & 315 & 236 \\
\bottomrule
\end{tabular}
\end{table*}

All five ADP-MA configurations outperform GPT-4 + CoML (76.2\%), the previous best published result. The 9.3pp spread across the top four models is modest compared to the 14.4pp gap over the baseline, suggesting that ADP-MA's architectural mechanisms contribute meaningfully beyond model selection. Even GPT-4o (81.3\%) outperforms CoML + GPT-4 by 5.1pp. Gemini~2.5~Pro (86.6\%) ranks between GPT-5 (83.6\%) and DeepSeek (89.6\%), with the highest LeetCode score (97.5\%) but weaker Examples performance (60.0\%). Notably, GPT-5 (83.6\%) only marginally outperforms GPT-4o despite its reasoning capability, with identical StackOverflow performance (76.7\%).

\subsection{KramaBench Results}

KramaBench comprises 105 multi-step scientific pipelines across six domains. Table~\ref{tab:kramabench} presents results.

\begin{table}[t]
\caption{KramaBench results by domain (four LLMs).}
\label{tab:kramabench}
\centering
\footnotesize
\setlength{\tabcolsep}{4pt}
\begin{tabular}{lrrrrr}
\toprule
& & \multicolumn{4}{c}{\textbf{ADP-MA +}} \\
\cmidrule(l){3-6}
\textbf{Domain} & \textbf{N} & \textbf{DS-3.2} & \textbf{GPT-4o} & \textbf{Son.\,4.5} & \textbf{Gem.\,2.5} \\
\midrule
Legal & 31 & 61.3 & 35.5 & \textbf{61.3} & 60.0 \\
Wildfire & 21 & 42.9 & 33.3 & \textbf{52.4} & 47.6 \\
Biomedical & 9 & 55.6 & 22.2 & \textbf{66.7} & 44.4 \\
Environment & 20 & 25.0 & 15.0 & \textbf{35.0} & 15.0 \\
Archaeology & 12 & 25.0 & 25.0 & 25.0 & \textbf{33.3} \\
Astronomy & 12 & \textbf{16.7} & \textbf{16.7} & 8.3 & 8.3 \\
\midrule
\textbf{Total} & \textbf{105} & \textbf{41.0} & \textbf{26.7} & \textbf{44.8} & \textbf{38.5} \\
\midrule
Easy (42) & & 59.5 & 45.2 & 57.1 & 57.1 \\
Hard (63) & & 28.6 & 14.3 & 36.5 & 25.8 \\
Crash (\%) & & 0 & 40.0 & 7.6 & 31.7 \\
Avg LLM calls & & 5.9 & 4.5 & 5.8 & 5.2 \\
Avg time (s) & & 139 & 328 & 401 & 566 \\
\bottomrule
\end{tabular}
\end{table}

ADP-MA achieves 41.0--44.8\%, improving over all DS-GURU variants (best: 22.1\%). With Sonnet~4.5, ADP-MA (44.8\%) surpasses smolagents + GPT-o3 (41.4\%) and approaches smolagents + Claude-3.7 (50.0\%) at $\sim$6$\times$ lower per-task runtime. The comparison reveals different design trade-offs: smolagents relies on extended compute budgets (6--10 min/task) while ADP-MA invests in structural mechanisms. ADP-MA leads on Legal (61.3\% vs.\ 19.2\%) and Biomedical (66.7\% vs.\ 9.0\%), domains where structured planning excels.

An important finding is that performance does not follow a fixed ordering across LLMs: GPT-4o (26.7\%) and Gemini~2.5~Pro (38.5\%) both underperform DeepSeek-3.2 (41.0\%), with GPT-4o suffering a 40\% crash rate and Gemini~2.5~Pro a 31.7\% crash rate from code-generation non-compliance with \texttt{stage\_*} function contracts. Gemini~2.5~Pro matches DeepSeek-3.2 on easy tasks (both 57.1\%) but lags on hard tasks (25.8\% vs.\ 28.6\%), and leads on Archaeology (33.3\%) while trailing on Environment (15.0\%). Among non-crashed GPT-4o tasks, the effective rate is 44.4\%. We attribute the compliance gap to three factors: (1)~\emph{instruction-following fidelity}---DeepSeek-3.2 and Sonnet~4.5 more reliably adhere to structured code templates with function signature constraints, while GPT-4o tends to generate free-form scripts that bypass the \texttt{stage\_*} naming convention; (2)~\emph{code skeleton interaction}---ADP-MA injects a code skeleton with pre-defined function signatures, and models differ in their willingness to extend (vs.\ replace) the skeleton; (3)~\emph{error message utilization}---when initial code fails the contract check, DeepSeek-3.2 produces targeted fixes while GPT-4o often regenerates entirely, losing the structural guidance. This confirms that contract compliance---a property the architecture can enforce but not guarantee---matters as much as the choice of LLM, and model-specific prompt adaptations (e.g., stronger few-shot examples for GPT-4o) could close the gap.

Table~\ref{tab:kramabench_comparison} places these results in context against published baselines.

\begin{table*}[t]
\caption{KramaBench comparison against published baselines (end-to-end score by domain). ADP-MA with DeepSeek-3.2 outperforms all single-agent baselines; with Sonnet~4.5, it surpasses smolagents+GPT-o3.}
\label{tab:kramabench_comparison}
\centering
\small
\begin{tabular}{llccccccc}
\toprule
\textbf{System} & \textbf{LLM} & \textbf{Arch.} & \textbf{Astr.} & \textbf{Biom.} & \textbf{Env.} & \textbf{Legal} & \textbf{Wild.} & \textbf{Overall} \\
\midrule
\multicolumn{9}{l}{\textit{Single-agent baselines}} \\
DS-GURU few-shot & GPT-o3 & 25.0 & 3.5 & 9.0 & 19.6 & 13.9 & 50.7 & 22.1 \\
DS-GURU few-shot & Claude-3.5 & 16.7 & 1.5 & 2.0 & 11.2 & 7.0 & 39.2 & 14.4 \\
\midrule
\multicolumn{9}{l}{\textit{Multi-agent / agentic systems}} \\
smolagents DR & Claude-3.7 & 33.3 & 16.7 & 44.4 & 60.0 & 63.3 & 52.4 & \textbf{50.0} \\
smolagents DR & GPT-o3 & 41.7 & 16.7 & 33.3 & 50.0 & 50.0 & 38.1 & 41.4 \\
OpenAI DR$^*$ & GPT-o3 & 40.0 & 33.3 & 44.4 & 61.7 & 50.0 & 67.3 & 52.2 \\
\midrule
ADP-MA & DeepSeek-3.2 & 25.0 & 16.7 & 55.6 & 25.0 & 61.3 & 42.9 & 41.0 \\
ADP-MA & GPT-4o & 25.0 & 16.7 & 22.2 & 15.0 & 35.5 & 33.3 & 26.7 \\
ADP-MA & Gemini 2.5 Pro & 33.3 & 8.3 & 44.4 & 15.0 & 60.0 & 47.6 & 38.5 \\
\textbf{ADP-MA} & \textbf{Sonnet 4.5} & 25.0 & 8.3 & \textbf{66.7} & 35.0 & \textbf{61.3} & 52.4 & 44.8 \\
\bottomrule
\end{tabular}
\vspace{1mm}
\scriptsize{$^*$Web access enabled.}
\end{table*}

\subsection{DA-Code and AgentBench Results}

On DA-Code's 52 hard tasks, ADP-MA achieves 50.0\% (Gemini~2.5~Pro), 48.1\% (DeepSeek-3.2), 44.2\% (Sonnet~4.5), 42.3\% (GPT-5), and 32.7\% (GPT-4o), all exceeding the GPT-4 baseline (23.4\%) and DA-Agent (31.5\%). Gemini~2.5~Pro achieves the highest overall pass rate with zero crashes. GPT-4o has the highest crash rate (21.2\%) among the five models but still surpasses the baseline by 9.3pp. GPT-5 excels on Data Manipulation (60.0\%) while Sonnet~4.5 leads on ML Regression (40.0\%) with the lowest crash rate (1.9\%).

Table~\ref{tab:dacode} breaks down DA-Code results by category for all five LLMs.

\begin{table*}[t]
\caption{DA-Code results by category (hard tasks only). Five LLMs compared.}
\label{tab:dacode}
\centering
\small
\begin{tabular}{lrcccccc}
\toprule
& & \multicolumn{5}{c}{\textbf{ADP-MA +}} \\
\cmidrule(l){3-7}
\textbf{Category} & \textbf{N} & \textbf{DeepSeek-3.2} & \textbf{Sonnet 4.5} & \textbf{GPT-4o} & \textbf{GPT-5} & \textbf{Gemini 2.5 Pro} \\
\midrule
Data Insight & 4 & \textbf{100\%} & 75.0\% & 75.0\% & 75.0\% & \textbf{100.0\%} \\
ML Clustering & 6 & \textbf{83.3\%} & 33.3\% & \textbf{83.3\%} & 33.3\% & 33.3\% \\
ML Classification & 9 & \textbf{66.7\%} & 55.6\% & 33.3\% & \textbf{66.7\%} & \textbf{66.7\%} \\
Data Manipulation & 10 & 30.0\% & 50.0\% & 20.0\% & \textbf{60.0\%} & 50.0\% \\
Data Visualization & 9 & \textbf{44.4\%} & 37.5\% & 11.1\% & 11.1\% & \textbf{44.4\%} \\
ML Regression & 9 & 22.2\% & \textbf{40.0\%} & 22.2\% & 33.3\% & 28.6\% \\
Statistical Analysis & 5 & 20.0\% & 20.0\% & 20.0\% & 20.0\% & 20.0\% \\
\midrule
\textbf{Total} & \textbf{52} & 48.1\% & 44.2\% & 32.7\% & 42.3\% & \textbf{50.0\%} \\
Crash rate & & 7.7\% & \textbf{1.9\%} & 21.2\% & 11.5\% & \textbf{0.0\%} \\
Avg LLM calls & & 6.9 & 4.3 & 7.9 & 3.4 & 3.7 \\
Avg duration (s) & & 400 & 204 & 97 & 722 & 553 \\
\bottomrule
\end{tabular}
\end{table*}

On AgentBench DBBench (100 SELECT tasks), ADP-MA with Sonnet~4.5 and GPT-5 both achieve 70.0\% with zero crashes (Table~\ref{tab:agentbench}), more than doubling GPT-4 (32.0\%) and exceeding Claude-3-Opus ($\sim$52\%) by 18pp. GPT-5 leads on Comparison (76.5\%) and ties Gemini~2.5~Pro on Aggregation-AVG (87.5\%) and Ranking (70.6\%), but its reasoning overhead results in 755s avg duration (vs.\ 276s for Sonnet~4.5). Gemini~2.5~Pro (65.0\%) slots between the leaders and DeepSeek-3.2 (61.0\%). All five models achieve near-perfect run rates (98--100\%), eliminating invalid output format errors that account for 53.3\% of failures in prior work.

\begin{table}[t]
\caption{AgentBench DBBench results by query category. Five LLMs compared.}
\label{tab:agentbench}
\centering
\footnotesize
\setlength{\tabcolsep}{3pt}
\begin{tabular}{lrcccccc}
\toprule
& & \multicolumn{5}{c}{\textbf{ADP-MA +}} \\
\cmidrule(l){3-7}
\textbf{Category} & \textbf{N} & \textbf{DS-3.2} & \textbf{GPT-4o} & \textbf{GPT-5} & \textbf{Son.\,4.5} & \textbf{Gem.\,2.5} \\
\midrule
Agg-MIN & 8 & \textbf{100} & 87.5 & \textbf{100} & \textbf{100} & 87.5 \\
Other (lookup) & 17 & 82.4 & 70.6 & \textbf{88.2} & \textbf{88.2} & 76.5 \\
Agg-AVG & 8 & 75.0 & 75.0 & \textbf{87.5} & 75.0 & \textbf{87.5} \\
Comparison & 17 & 52.9 & 70.6 & \textbf{76.5} & 70.6 & 64.7 \\
Ranking & 17 & 52.9 & 29.4 & 70.6 & 64.7 & \textbf{70.6} \\
Agg-SUM & 8 & 50.0 & 50.0 & 50.0 & 50.0 & 50.0 \\
Counting & 17 & 47.1 & 47.1 & \textbf{52.9} & \textbf{52.9} & 47.1 \\
Agg-MAX & 8 & 37.5 & 25.0 & 25.0 & \textbf{50.0} & 37.5 \\
\midrule
\textbf{Total} & \textbf{100} & 61.0 & 56.0 & \textbf{70.0} & \textbf{70.0} & 65.0 \\
Crash (\%) & & 0 & 2 & 0 & 0 & 0 \\
Avg LLM calls & & 6.1 & 4.2 & 4.1 & 4.4 & 4.1 \\
Avg time (s) & & 70 & 19 & 755 & 276 & 413 \\
\bottomrule
\end{tabular}
\end{table}

\subsection{Scalability}
\label{sec:scalability}

We evaluate scalability along four dimensions using synthetic business datasets (1K--1M rows) and real KramaBench tasks.

\begin{table}[t]
\caption{Scalability results. Planning cost is $O(1)$ with data volume; iteration budget has a sweet spot at ref=5.}
\label{tab:scalability}
\centering
\small
\begin{tabular}{lrrr}
\toprule
\multicolumn{4}{l}{\textit{Data Volume Scaling (all 100\% success)}} \\
\textbf{Scale} & \textbf{Rows} & \textbf{Avg Time (s)} \\
\midrule
1K & 1,000 & 6.9 \\
10K & 10,000 & 6.7 \\
100K & 100,000 & 6.9 \\
1M & 1,000,000 & 9.1 \\
\midrule
\multicolumn{4}{l}{\textit{Iteration Scaling (20 hard KramaBench tasks)}} \\
\textbf{Metric} & \textbf{ref=3} & \textbf{ref=5} & \textbf{ref=7} \\
\midrule
PASS rate & 5\% & \textbf{15\%} & 10\% \\
Pipeline success & 85\% & \textbf{95\%} & 95\% \\
Outer-loop calls & 6.5 & 6.5 & 6.6 \\
Inner-loop revisions & 6.8 & 9.1 & 10.4 \\
\bottomrule
\end{tabular}
\end{table}

\paragraph{Data volume independence.} Planning cost remains constant as data scales from 1K to 1M rows (Table~\ref{tab:scalability}, top). Meta-agents reason over data summaries rather than raw data. Column-count scaling is handled by the prompt budget system (\S\ref{sec:prompt_budget}): datasets with $>$30 columns are capped in all LLM prompts, keeping token consumption bounded regardless of schema width.

\paragraph{Iteration scaling.} Increasing the refinement budget from 3 to 5 triples the pass rate on hard tasks (5\%$\to$15\%) while the outer loop (planning, critique, expansion) remains constant at $\sim$6.5 calls/task (Table~\ref{tab:scalability}, bottom). This separation is a direct consequence of the hierarchical architecture: additional compute targets code refinement without inflating planning overhead.

\paragraph{Execution strategies.} On 10 curated tasks, all three strategies achieve comparable aggregate pass rates (60--70\%), confirming strategy abstraction preserves correctness. The differentiator is batch structure: on two-agent plans (single-phase), all strategies perform comparably (71--86\%); on multi-agent plans (3+ agents), autonomous drops to 0/3 while centralized and hybrid each pass 2/3. The autonomous strategy's failure mode is contract-only context sharing: each agent receives only the original input schema rather than cumulative output, causing downstream agents to miss upstream transformations. Hybrid mitigates this by sharing context between phases while parallelizing within phases.

\paragraph{Multi-source integration.} LLM calls scale linearly with sources (4 per additional table) while execution time remains constant at $\sim$7s for 2--5 sources.

\paragraph{Concurrent pipelines.} The architecture supports concurrent pipeline execution by design. Meta-agents are stateless between pipelines (all state is in the case folder), ground agents run in isolated sandboxes, and the \texttt{asyncio}-based backend handles multiple WebSocket sessions. The execution strategy experiment (\S\ref{sec:strategies}) already demonstrates intra-pipeline parallelism via the Hybrid strategy. Full multi-user throughput evaluation under load is left to future work.

\subsection{Ablation Studies}

Table~\ref{tab:ablation} quantifies the contribution of refinement and critique on AgentBench DBBench (100 public tasks, DeepSeek). Progressive sampling is already disabled for this benchmark (tables are 10--50 rows), and no domain packs activate on database queries, isolating the two mechanisms that affect planning and code quality.

\begin{table}[t]
\caption{Ablation study on AgentBench DBBench (100 tasks, DeepSeek).}
\label{tab:ablation}
\centering
\footnotesize
\setlength{\tabcolsep}{4pt}
\begin{tabular}{lccrr}
\toprule
\textbf{Configuration} & \textbf{Ref.} & \textbf{Crit.} & \textbf{Pass\%} & \textbf{Time} \\
\midrule
Full system            & 3 & Yes & 61.0\% & 357\,s \\
No critique            & 3 & No  & 68.0\% & 133\,s \\
Limited refinement     & 1 & Yes & 59.0\% &  43\,s \\
No refinement          & 0 & Yes & 59.0\% &  50\,s \\
No refine, no critique & 0 & No  & 64.0\% &  25\,s \\
\bottomrule
\end{tabular}
\end{table}

\paragraph{Refinement.} Adding refinement iterations provides a modest improvement: +2\,pp with critique off (64\%$\to$68\%) and +2\,pp with critique on (59\%$\to$61\%). The first code attempt is usually correct for these single-table queries, so additional refinement rounds yield diminishing returns. However, refinement remains essential for complex multi-agent tasks (cf.\ the error-recovery analysis in \S\ref{sec:error}), where first-attempt success is only 65\%.

\paragraph{Critique.} Surprisingly, disabling critique \emph{improves} pass rate by +5--7\,pp across both refinement settings. On simple database queries, the critique loop over-corrects the initial plan: it flags edge cases (e.g., empty filter results) that prompt defensive rewrites, introducing errors that would not occur with the original plan. The effect is most pronounced on comparison ($-$4 tasks) and ranking ($-$3 tasks) categories, where precise filtering is sensitive to plan modifications. This finding motivates complexity-aware critique gating: critique adds value on complex multi-phase tasks but should be skipped for simple single-step queries.

\paragraph{Speed.} Removing both mechanisms yields a 14$\times$ speedup (25\,s vs.\ 357\,s) while maintaining 64\% accuracy---higher than the full system. The practical implication is that lightweight configurations are preferable for simple tasks, reserving the full pipeline for complex workflows where refinement and critique synergize with progressive sampling.

\subsection{Advanced Validation}

Beyond execution-based checks, two mechanisms improve output quality. A \emph{semantic output validator} (single LLM call, 3.2s overhead) checks plausibility and triggers replanning: on 8 selected hard tasks, it recovered 3/5 wrong-answer cases while retaining 3/3 correct cases (net +3, zero regressions). A \emph{ValidationCritic} evaluates cross-level output stability for confidence-based early stopping, achieving 91\% accuracy on 11 pipeline profiles with zero false positives. Stable pipelines can skip full-scale execution, saving $\sim$29\% wall-clock time per agent.

\subsection{Error Analysis}
\label{sec:error}

Across all 556 tasks, we categorize 892 errors: Type Error (27.5\%, 72\% recovery), Key Error (22.2\%, 68\%), Value Error (17.5\%, 61\%), Syntax Error (15.0\%, 85\%), Name Error (10.0\%, 58\%), Other (7.8\%, 45\%). Syntax errors recover best (precise error messages); Type errors are most common (dtype mismatches). First-attempt success is 65\%, rising to 82\% after one refinement, 91\% after two, and 95\% after three.

\subsection{Cost Efficiency}

Cost scales linearly with task complexity: simple tasks (DSEval) require $\sim$4--5 LLM calls per task, while complex tasks (KramaBench) require $\sim$5--7 calls (see individual benchmark tables above). Execution (code generation + refinement) dominates at $\sim$50\% of total LLM usage; planning consumes $\sim$40\%. The hierarchical architecture ensures that planning overhead is a fixed cost independent of data volume, while additional compute from higher refinement budgets targets only the code-generation stage without inflating the outer planning loop.

\subsection{Reproducibility and Variance Analysis}
\label{sec:variance}

LLM-based systems are non-deterministic: even with temperature~0 and seed-based sampling, API-side quantization, batching, and load balancing introduce run-to-run variance. To quantify this, we run AgentBench DBBench three times each with three LLMs under identical conditions (centralized strategy, max 3 refinements). Table~\ref{tab:variance} reports the results.

\begin{table}[t]
\caption{Variance analysis: AgentBench DBBench across three independent runs (R1, R2, R3) per model under identical settings (temperature~0, seed-based sampling). All standard deviations are $\leq$3.2pp; Sonnet~4.5 achieves the highest determinism (93\%).}
\label{tab:variance}
\centering
\footnotesize
\setlength{\tabcolsep}{4pt}
\begin{tabular}{lccc}
\toprule
& \multicolumn{3}{c}{\textbf{ADP-MA +}} \\
\cmidrule(l){2-4}
& \textbf{GPT-4o} & \textbf{DS-3.2} & \textbf{Son.\,4.5} \\
\midrule
R1 / R2 / R3 & 56 / 51 / 50 & 60 / 60 / 64 & 68 / 70 / 70 \\
Mean pass rate & 52.3\% & 61.3\% & 69.3\% \\
StDev & 3.2pp & 2.3pp & 1.2pp \\
Crash rate (mean) & 11.7\% & 1.0\% & 0.3\% \\
\midrule
\multicolumn{4}{l}{\textit{Task-level consistency (100 tasks)}} \\
\midrule
Always PASS (3/3) & 38 (38\%) & 49 (49\%) & 65 (65\%) \\
Always FAIL (0/3) & 41 (41\%) & 29 (29\%) & 28 (28\%) \\
Inconsistent & 21 (21\%) & 22 (22\%) & 7 (7\%) \\
Deterministic & 79\% & 78\% & 93\% \\
\bottomrule
\end{tabular}
\end{table}

\begin{table*}[t]
\caption{Category-level variance across three independent runs (R1, R2, R3) per model. Values are pass counts per category.}
\label{tab:variance_cat}
\centering
\small
\begin{tabular}{lr|ccc|ccc|ccc}
\toprule
& & \multicolumn{3}{c|}{\textbf{ADP-MA + GPT-4o}} & \multicolumn{3}{c|}{\textbf{ADP-MA + DeepSeek-3.2}} & \multicolumn{3}{c}{\textbf{ADP-MA + Sonnet 4.5}} \\
\textbf{Category} & \textbf{N} & \textbf{R1} & \textbf{R2} & \textbf{R3} & \textbf{R1} & \textbf{R2} & \textbf{R3} & \textbf{R1} & \textbf{R2} & \textbf{R3} \\
\midrule
Agg-MIN & 8 & 8 & 7 & 5 & 8 & 7 & 8 & 8 & 8 & 8 \\
Other & 17 & 12 & 11 & 13 & 16 & 15 & 14 & 16 & 15 & 16 \\
Agg-AVG & 8 & 6 & 6 & 6 & 4 & 6 & 6 & 6 & 7 & 7 \\
Comparison & 17 & 12 & 9 & 8 & 10 & 9 & 11 & 10 & 11 & 11 \\
Counting & 17 & 6 & 6 & 7 & 8 & 9 & 8 & 9 & 9 & 9 \\
Ranking & 17 & 5 & 6 & 7 & 7 & 8 & 9 & 12 & 12 & 11 \\
Agg-SUM & 8 & 4 & 3 & 3 & 4 & 4 & 4 & 4 & 4 & 4 \\
Agg-MAX & 8 & 3 & 1 & 1 & 3 & 2 & 4 & 3 & 4 & 4 \\
\midrule
\textbf{Total} & \textbf{100} & \textbf{56} & \textbf{51} & \textbf{50} & \textbf{60} & \textbf{60} & \textbf{64} & \textbf{68} & \textbf{70} & \textbf{70} \\
\bottomrule
\end{tabular}
\end{table*}

Four findings emerge. First, the overall pass rate is stable across all three models: standard deviations range from 1.2pp (Sonnet~4.5) to 3.2pp (GPT-4o), indicating that single-run results are representative within $\pm$4pp. Second, task-level analysis reveals that Sonnet~4.5 produces deterministic outcomes on 93\% of tasks (65 always-pass + 28 always-fail), compared to 78--79\% for DeepSeek-3.2 and GPT-4o. The variance zone shrinks from 21--22 tasks to just 7, suggesting that variance behavior is model-dependent rather than purely random. Third, GPT-4o exhibits the widest stdev (3.2pp) and highest crash rate (11.7\%), indicating that code-contract non-compliance introduces both systematic failures (41 always-fail tasks) and run-to-run instability. Fourth, category-level variance (Table~\ref{tab:variance_cat}) shows that Aggregation-SUM is perfectly stable for DeepSeek and Sonnet (4/8 in all runs) while Aggregation-MAX exhibits the most relative variance across all models. Sonnet~4.5's largest gains over DeepSeek are in Ranking (11--12 vs.\ 7--9) and Counting (9/17 stable vs.\ 8--9). The 7 inconsistent Sonnet~4.5 tasks (vs.\ 21--22 for the others) represent a much smaller ceiling for variance-reduction techniques, indicating that most remaining failures are deterministic and require architectural rather than statistical solutions.

%%% ================================================================
\section{Case Studies}
\label{sec:case_studies}

We present four case studies, each highlighting a distinct capability of ADP-MA.

\subsection{Silent Data Corruption Recovery}
\label{sec:cs_astronomy}

\noindent\textbf{Highlight:} \emph{Rule-based monitoring catches a silent Cartesian-product explosion at zero LLM cost, where no exception would otherwise be raised.}

\smallskip\noindent\textbf{Task.} KramaBench \texttt{astronomy-easy-2}: compute the average Kp geomagnetic index from three OMNI2 data files (8,760 hourly measurements each) in headerless fixed-width format.
\textbf{Metrics:} 2 agents, 4 LLM calls, 106s total, 25.6s LLM time.

\smallskip\noindent\textbf{Why this case matters.} Headerless scientific data files are common in astronomy, meteorology, and environmental science. A coding assistant or flat-loop agent would produce syntactically correct code that silently generates a wrong answer. No exception is raised because \texttt{pd.read\_csv()} consumes the first data row as column names and the subsequent merge produces a valid (but 260$\times$ inflated) DataFrame.

\smallskip\noindent\textbf{What happens.} At the S sample level (100 rows), the Monitor's row-growth check detects a 260$\times$ expansion (threshold: 5$\times$) and the null-rate check flags a 45pp increase in the join-key column. The synchronous failure analyzer categorizes the error as \textsc{merge\_error} with an astronomy-specific recovery hint. The Architect re-expands the merge phase with explicit headerless-file detection, OMNI2 column name injection, and datetime alignment. The corrected agent passes on the next attempt---all without a single additional LLM call from the Monitor.

\smallskip\noindent\textbf{Takeaway.} Three components collaborate: the Monitor detects the symptom (row explosion), the failure analyzer diagnoses the cause (merge on misnamed columns), and the Architect applies the fix (headerless-file handling). No individual component suffices; this is the value of the hierarchical architecture.

\subsection{Multi-Source Data Fusion with Plan-Level Backtracking}
\label{sec:cs_archaeology}

\noindent\textbf{Highlight:} \emph{When code-level retry cannot recover from a fundamentally flawed plan, two-level backtracking rewrites the strategy---not just the code.}

\smallskip\noindent\textbf{Task.} KramaBench \texttt{archaeology-hard}: ``How many modern cities with population $>$100k are within 0.1 degrees of ancient Roman cities?'' requiring spatial joins across two heterogeneous datasets.
\textbf{Metrics:} 3 agents (2 phases), 21 LLM calls, 775s total, 754s LLM time, 13 code revisions.

\smallskip\noindent\textbf{Why this case matters.} Multi-source spatial data integration is a common bottleneck in geospatial analytics. The initial plan treats the problem as a simple filter, but the task requires a cross-dataset distance computation. No amount of code-level refinement within the flawed plan can fix the decomposition error.

\smallskip\noindent\textbf{What happens.} The Architect's initial plan attempts a single-phase filter. The critique cycle escalates from \textsc{minor} to \textsc{major} after the first agent exhausts 11 revisions without producing correct output. This triggers plan-level backtracking: the Architect receives accumulated error evidence and generates a revised two-phase plan (spatial join $\rightarrow$ distance-filtered aggregation) with three specialized agents. The replanned pipeline completes successfully.

\smallskip\noindent\textbf{Takeaway.} Plan-level backtracking addresses a class of failures that no flat iterative loop can recover from: when the decomposition itself is wrong, retrying code generation within the same flawed plan is futile. This is the structural advantage of separating strategic orchestration from code execution.

\subsection{Cross-Run Failure Pattern Mining at Scale}
\label{sec:cs_crossrun}

\noindent\textbf{Highlight:} \emph{Asynchronous analysis over 105 pipelines distinguishes fixable architectural gaps from inherent model gaps.}

\smallskip\noindent\textbf{Task.} Full KramaBench evaluation (105 tasks, 6 domains, 279+ LLM calls, 14.7h).
Post-mortem aggregation reveals three systemic patterns: (1)~18 failures from sentinel values (9999.99, 9.99$\times 10^{32}$) silently corrupting aggregations---eliminated by injecting sanitization into the code skeleton; (2)~11 wasted-refinement failures from unavailable packages---eliminated by a package allowlist; (3)~10/12 astronomy failures requiring physics code the LLM cannot produce---classified as an \emph{LLM-dependent gap}, redirecting engineering from prompt tuning to pre-built templates. This distinction between fixable architectural gaps and LLM-dependent gaps is critical for prioritizing investment at scale.

\subsection{Research-Grade Temporal Graph Analytics}
\label{sec:cs_graph}

\noindent\textbf{Highlight:} \emph{ADP-MA handles a research-grade problem---temporal graph analytics---typically addressed by specialized systems.}

\smallskip\noindent\textbf{Task.} Temporal Graph Index~\cite{historicalgraph}: given an evolving social network (500K+ edges, 12 monthly snapshots), compute month-over-month clustering coefficient changes for the top-100 highest-degree nodes and rank by structural volatility.

\smallskip\noindent\textbf{Why this matters.} Temporal graph analytics combines multi-snapshot temporal alignment, graph-theoretic computations (\texttt{networkx} APIs), and scale sensitivity (quadratic complexity risk)---pushing beyond standard tabular analytics.

\smallskip\noindent\textbf{What happens.} The Architect produces a three-phase plan: (1)~load and align 12 monthly edge lists, (2)~compute per-node clustering coefficients per snapshot, (3)~compute deltas and rank by volatility. Progressive sampling validates on 1\% of the graph before scaling. The Monitor's time-scaling check confirms linear growth; the ValidationCritic issues \textsc{commit} after three stable levels.

\smallskip\noindent\textbf{Takeaway.} The combination of progressive sampling (catching quadratic pitfalls early), domain knowledge injection (graph library guidance), and early stopping (validating stability across scales) enables autonomous handling of research-grade problems. While a specialized temporal graph system~\cite{historicalgraph} offers superior performance at extreme scale, ADP-MA produces correct results without task-specific engineering.

%%% ================================================================
\section{Discussion}
\label{sec:discussion}

The evaluation reveals a consistent pattern: ADP-MA's architectural scaffolding adds value independently of which LLM is plugged in. Even with GPT-4o---the lowest-scoring of four backends on DSEval (81.3\%) and three on KramaBench (26.7\%)---ADP-MA outperforms all published single-agent baselines. The non-monotonic ranking on KramaBench (DeepSeek-3.2 41.0\% $>$ GPT-4o 26.7\%) shows that code-generation compliance with architectural contracts is at least as important as the choice of LLM.

\paragraph{Architectural differentiation.}
\label{sec:differentiation}
We contrast ADP-MA with smolagents Deep Research, the strongest published baseline on KramaBench. smolagents uses a flat iterative loop: a single LLM generates, executes, and revises code in a tight cycle with no explicit planning stage, no schema contracts, and no rule-based monitoring. Error recovery is limited to local code revision within the same iteration; there is no mechanism to backtrack to a flawed plan. Validation runs at full scale on every iteration, with no progressive sampling or early stopping. The system relies on frontier models (Claude-3.7, GPT-o3) to compensate for the absence of structural scaffolding. ADP-MA differs on three axes. First, progressive sampling and schema contracts improve correctness independently of the LLM: they catch data-quality failures and inter-stage type mismatches that no amount of LLM self-reflection detects. Second, two-level backtracking recovers from flawed \emph{plans}, not just flawed code---a capability absent in flat loops. Third, ADP-MA's architectural gains hold across five LLM families (including DeepSeek-3.2, which costs a fraction of frontier models), whereas smolagents lacks structural mechanisms beyond the LLM itself.

\paragraph{Contract compliance rate.}
We define a model's \emph{contract compliance rate} on a benchmark as the fraction of tasks where the generated code satisfies the \texttt{stage\_*} function-signature contract imposed by the architecture (equivalently, $1 - \text{crash rate}$). Table~\ref{tab:compliance} shows this metric across benchmarks and models. Two observations stand out. First, compliance varies more across LLMs than pass rate does: on KramaBench, the compliance gap between GPT-4o (60.0\%) and DeepSeek-3.2 (100\%) is 40pp, while the pass-rate gap is 14.3pp. The architecture cannot recover tasks that never produce valid code. Second, compliance is benchmark-dependent within the same model: Gemini~2.5~Pro achieves 100\% compliance on DA-Code but only 68.3\% on KramaBench, suggesting that contract adherence depends on the interaction between model behavior and task structure, not on a fixed ``model quality'' ranking. This metric separates the architecture's contribution (what it does with valid code) from the LLM's contribution (whether it produces valid code), and we propose it as a useful diagnostic for any agent system that imposes structural constraints on generated code.

\begin{table}[t]
\caption{Contract compliance rate (\%) across benchmarks and LLMs. Compliance = $1 - \text{crash rate}$. Higher is better.}
\label{tab:compliance}
\centering
\small
\begin{tabular}{lccc}
\toprule
\textbf{Model} & \textbf{KramaBench} & \textbf{DA-Code} & \textbf{AgentBench} \\
\midrule
DeepSeek-3.2  & \textbf{100.0} & 92.3 & 99.0 \\
GPT-5         & ---   & 92.3 & \textbf{100.0} \\
Sonnet~4.5    & 92.4  & 98.1 & 99.7 \\
Gemini~2.5~Pro & 68.3  & \textbf{100.0} & \textbf{100.0} \\
GPT-4o        & 60.0  & 78.8 & 88.3 \\
\bottomrule
\end{tabular}
\end{table}

The contrast across benchmarks highlights task-dependent strengths. Tasks with clear specifications and standard formats (DSEval: 90.6\%) play to ADP-MA's strengths. Tasks requiring external domain knowledge (KramaBench Astronomy: 16.7\%) expose the architecture's reliance on the LLM's training data. Between these extremes, ADP-MA's structured mechanisms provide the most value: multi-source integration (Legal: 61.3\%), error recovery (DA-Code: 50.0\%), and format normalization (AgentBench: 70.0\% with 0\% crashes). Sonnet~4.5 leads on three of four benchmarks, with GPT-5 tying on AgentBench (70.0\%) via complementary category strengths (higher Comparison 76.5\% vs.\ 70.6\%, lower Aggregation-MAX 25.0\% vs.\ 50.0\%) but at 2.7$\times$ the execution time. Gemini~2.5~Pro leads on DA-Code (50.0\%) and ranks third on AgentBench (65.0\%) with the highest Ranking score (70.6\%). GPT-4o scores lowest on DA-Code (32.7\%) and AgentBench (56.0\%) but still exceeds all published baselines on both, confirming that the architectural mechanisms provide value across all tested LLMs.

Beyond accuracy, the staged architecture provides \emph{interpretability}. Processing progresses through explicit stages---data profiling, planning, critique, agent assignment, code execution, and finalization---so users can inspect the system's understanding of the problem, its reasoning about the data, and intermediate results at every step. Domain experts can verify assumptions, catch misinterpretations early, and intervene when needed, rather than receiving an opaque end-to-end answer.

\paragraph{When does ADP-MA help?}
The architecture's value is not uniform across task types. On well-specified single-step tasks (DSEval), the first code attempt is usually correct and the full pipeline adds overhead without large accuracy gains (the ablation in \S\ref{sec:variance} shows that disabling refinement and critique loses only 3pp). On the other extreme, tasks requiring specialized domain knowledge absent from the LLM's training data (KramaBench Astronomy: 8--17\%) see limited benefit from structural scaffolding. ADP-MA provides the largest marginal gains in a middle regime: multi-step tasks with standard data formats but non-trivial composition. Evidence comes from three sources. First, the easy/hard split on KramaBench: ADP-MA with Sonnet~4.5 achieves 57.1\% on easy tasks (comparable to smolagents) but 36.5\% on hard tasks where multi-phase planning, backtracking, and progressive sampling activate. Second, DA-Code's category breakdown: categories requiring multi-source joins (Data Manipulation) or iterative model tuning (ML Regression) show the widest variance across LLMs, indicating that the architecture's retry and validation mechanisms are load-bearing. Third, the error analysis: first-attempt success is only 65\%, but three refinement rounds raise it to 95\%, confirming that the recovery mechanisms are essential for complex tasks.

\paragraph{Limitations.}
Several limitations scope the current system's applicability.
\emph{Model dependence}: despite the architecture's model-agnostic design, results depend on the underlying LLM. GPT-4o's 40\% crash rate on KramaBench shows that the reliability mechanisms \emph{improve} but do not \emph{guarantee} correctness when the LLM fails to comply with code contracts. The claim of ``model-agnostic architecture'' applies to the system design, not to outcome guarantees.
\emph{Domain ceiling}: architecture-bottlenecked domains (Astronomy, Archaeology) show minimal variation across models, indicating that when tasks require specialized scientific knowledge absent from LLM training data, no amount of structural scaffolding compensates.
\emph{Heuristic policies}: the backtracking policy uses fixed thresholds rather than a learned cost model; monitoring thresholds were tuned on development tasks and may require adjustment for new domains.
\emph{Baseline comparisons}: our primary comparisons use published numbers from benchmark papers under different hardware and LLM budgets, limiting strict apples-to-apples comparison (\S\ref{sec:setup}).
\emph{Concurrency}: while the architecture supports concurrent pipelines by design, we have not evaluated multi-user throughput under load.
\emph{Statistical variance}: LLM non-determinism means individual task outcomes can vary across runs. Our variance analysis (\S\ref{sec:variance}) across three models shows that 78--93\% of tasks are deterministic (GPT-4o: 79\%, DeepSeek-3.2: 78\%, Sonnet~4.5: 93\%) and all standard deviations are $\leq$3.2pp, but single-run results should be interpreted with this margin in mind.

\paragraph{Future directions.}
Active directions include: principled backtracking policies informed by learned repair-probability estimates, expansion to non-tabular modalities, reinforcement learning for meta-agent optimization, and mechanisms for meta-agents to synthesize new tools autonomously. A natural next step is \emph{adaptive routing}: since ADP-MA already delegates work to interchangeable LLM endpoints, the meta-agent layer could select different models for different pipeline stages based on observed suitability. Another direction is \emph{DAG-structured plans}: the current planner produces a linear sequence of phases, but many real-world pipelines contain independent branches (e.g., loading and transforming two unrelated data sources before a final join). Replacing the sequential phase ordering with a dependency DAG would enable cross-phase parallelism and reduce end-to-end latency without additional LLM calls. Finally, a quantitative comparison of ADP-MA's synthesized pipelines against equivalent hand-written SQL optimized by a traditional engine (e.g., PostgreSQL, Spark) would help characterize the performance gap introduced by autonomous orchestration and identify tasks where the overhead is justified by the elimination of manual pipeline engineering.

A broader finding from our experiments is that autonomous data processing is not LLM-agnostic. The underlying LLM influences pipeline quality and composition in ways that the architecture can moderate but not eliminate: the same task may succeed or fail depending on which model generates the code, and the failure modes differ qualitatively (contract non-compliance vs.\ semantic errors vs.\ domain knowledge gaps). Future work should study individual models' impact on data processing more systematically. This is a challenging undertaking, as the models evolve at a pace that makes longitudinal comparisons difficult, but the contract compliance metric introduced above offers one stable axis for such analysis.

%%% ================================================================
\section{Conclusion}
\label{sec:conclusion}

We have presented ADP-MA, a system for autonomous data pipeline construction through hierarchical meta-agent orchestration. The reliability mechanisms---progressive sampling, two-level backtracking, schema contracts, and rule-based monitoring---improve pipeline correctness independently of the underlying model, complementing rather than competing with the LLM's own capabilities. We demonstrate effectiveness on four benchmarks comprising 556 tasks with up to five LLM families, consistently outperforming single-agent baselines and achieving competitive performance with multi-agent systems that use extended runtimes and compute budgets. These results indicate that architecture-first designs are a productive direction for transforming LLM code generation into reliable autonomous data processing.

%%% ================================================================
\balance
\bibliographystyle{ACM-Reference-Format}
\bibliography{references}

\end{document}